\documentclass[times, review, 10pt]{elsarticle}

\makeatletter
\usepackage{setspace}
\makeatletter
\renewenvironment{abstract}{\global\setbox\absbox=\vbox\bgroup
  \hsize=\textwidth
  \noindent\unskip\textbf{\@elsarticleabstitle}
 \par\medskip\noindent\unskip\ignorespaces}
 {\egroup}
\makeatother

\makeatother
\usepackage{amsmath}
\usepackage{amssymb}
\usepackage{amsfonts}
\usepackage{booktabs}
\usepackage{makecell}
\usepackage{multirow}
\usepackage{array}
\usepackage{graphicx}
\usepackage{mathrsfs}
\usepackage{textcomp}
\usepackage{tabularx}
\usepackage{url}
\usepackage{threeparttable}

\biboptions{numbers,sort&compress}
\journal{Pattern Recognition}

\usepackage{longtable}
\usepackage{booktabs}
\usepackage{makecell}
\newcommand{\prfloatbody}{\rmfamily\fontsize{10pt}{12pt}\selectfont}
\newcommand{\prtablebody}{\rmfamily\fontsize{8pt}{16pt}\selectfont\singlespacing}
\renewcommand{\footnotesize}{\fontsize{8pt}{10pt}\selectfont}
\renewcommand{\pm}{\mathbin{\text{\textpm}}}
\newcommand{\rs}[2]{#1\,\ensuremath{\pm}\,#2}
\newcommand{\rsb}[2]{\textbf{#1\,\ensuremath{\pm}\,#2}}
\newcommand{\rsu}[2]{\underline{#1\,\ensuremath{\pm}\,#2}}

\begin{document}
\pagenumbering{arabic}
\pagestyle{plain}
\renewcommand{\topfraction}{0.9}
\renewcommand{\bottomfraction}{0.7}
\renewcommand{\textfraction}{0.05}
\renewcommand{\floatpagefraction}{0.5}
\setcounter{topnumber}{4}
\setcounter{bottomnumber}{2}
\setcounter{totalnumber}{6}

\begin{frontmatter}

\title{A Pathway-Structured Privileged Distillation Framework for Deployable Computational Pathology}

\author[wfusm]{Yongxin Guo\corref{cor1}}
\ead{Yongxin.Guo@wfusm.edu}

\author[wfusm]{Hao Lu}
\author[wfusm]{Onur C. Koyun}
\author[wfusm]{Muhammet F. Demir}
\author[osu]{M Khalid Khan Niazi}
\author[wfusm]{Metin N. Gurcan}

\affiliation[wfusm]{organization={Wake Forest University School of Medicine},
            city={Winston-Salem},
            state={NC},
            country={USA}}
\affiliation[osu]{organization={Department of Pathology, The Ohio State University Wexner Medical Center},
            city={Columbus},
            state={OH},
            country={USA}}

\cortext[cor1]{Corresponding author.}

\begin{abstract}
Integrating transcriptomics and histopathology can improve cancer risk modelling, yet clinical translation is constrained by the limited availability of RNA profiling in routine practice. Here, we introduce Mixture of Pathway Experts (MoPE), a pathway-structured privileged-distillation framework for histology-only inference. MoPE is motivated by the partial correspondence between transcriptomic state and tissue morphology: histology may capture morphological consequences of selected molecular programmes but cannot be expected to recover the full transcriptomic state. To accommodate this incomplete correspondence, MoPE organizes RNA-derived supervision through biological pathways and transfers it to pathway-indexed histology experts via memory-usage distillation. Across diverse public benchmarks and two independent external breast cancer cohorts, comprising 5,786 whole-slide images in total, MoPE consistently improved WSI-only biomarker classification and survival prediction compared with baseline methods. Pathway-usage analyses and human expert inspection further enabled structured examination of model behaviour and morphology-linked pathway hypotheses. Together, these results support pathway-structured privileged distillation as a practical strategy for exploiting molecular information during training while retaining RNA-free inference especially under limited clinical scenarios.
\end{abstract}


\end{frontmatter}

\section{Introduction}\label{sec:intro}

Whole-slide images (WSIs) are routinely generated during cancer diagnosis and provide a scalable record of tumour morphology, tissue architecture \citep{campanella2019clinical} and microenvironmental context \citep{hanahan2011hallmarks}. In parallel, molecular assays, including RNA-based signatures and genomic biomarkers, increasingly refine prognostic and therapeutic stratification; the 21-gene recurrence score in breast cancer is a representative example of how RNA expression-based measurements can guide risk assessment and adjuvant treatment decisions \citep{paik2004multigene,sparano2018adjuvant,howard2023integration}. However, RNA profiling and broader clinical sequencing are not consistently available in routine care because they require sufficient tissue \citep{byron2016translating}, validated laboratory workflows \citep{damodaran2015clinical}, additional cost and turnaround times \citep{su2025computational}. This creates a \textit{deployment asymmetry}: paired histology--molecular cohorts can support model development, whereas many real-world settings require inference from H\&E images alone. 

This asymmetry is compounded by \textit{partial observability}. Clinically validated expression assays show that RNA profiles carry prognostic and treatment-relevant information that is not fully captured by routine clinicopathological assessment \citep{sparano2018adjuvant}. H\&E morphology is therefore not a second measurement of the RNA. Consistent with this view, computational pathology studies have shown that selected mutations, molecular subtypes and gene-expression signatures can be inferred from routine H\&E images, but only through morphology-accessible correlates rather than direct molecular measurement \citep{kather2020pan}(Fig.~\ref{fig:model_overview}a). Moreover, bulk RNA profiles aggregate signals across heterogeneous cell populations, whereas histology preserves spatial organization at cellular resolution. Molecular supervision for histology-only deployment should therefore exploit this partial correspondence. The goal is not to reconstruct molecular profiles from images, but to transfer biologically meaningful structure that is reliably expressed through tissue morphology while preserving modality-specific information that remains unobservable in H\&E sections.

Existing computational pathology approaches address parts of this problem, but not the full dual constraint of H\&E-only deployment under incomplete molecular observability. In particular, most distillation strategies transfer molecular information through direct feature matching, teacher--student similarity preservation \citep{xing2024comprehensive} or subspace alignment \citep{zhang2026disentangled}, which can implicitly treat molecular representations as fully transferable to histology. These approaches leave a central question: how can molecular supervision be transferred to histology-only models when histology provides only a partial observability of molecular state? The representation used for distillation is therefore crucial. Distilling the large gene bank ($n>$20,000) creates a high-dimensional target. Biological pathways provide a more suitable unit of supervision \citep{liberzon2015molecular}: they group genes into compact and interpretable programmes which linked to tissue-level phenotypes, including proliferation, and stromal remodelling. Pathway-level supervision can therefore organize RNA-derived information in a biologically meaningful form.

Here we introduce Mixture of Pathway Experts (MoPE), a pathway-structured privileged-distillation framework for histology-only computational pathology (Fig.~\ref{fig:model_overview}). During training, paired RNA profiles are encoded as pathway tokens, and the WSI branch uses pathway-indexed experts to process image features. MoPE transfers privileged information through memory-usage distillation: RNA pathway tokens and pathway-indexed experts are encouraged to use a shared memory basis in similar ways, rather than being forced into direct feature identity. At inference, the RNA branch is removed and predictions are made from WSIs alone. We evaluated MoPE across five biomarker classification and five survival prediction tasks from The Cancer Genome Atlas (TCGA: \url{https://www.cancer.gov/tcga}), and further validated its generalizability on two independent clinical cohorts.

The main contributions of this work are summarized as follows.
\begin{itemize}
\item We reframe molecularly supervised histology-only prediction as a \textit{partial-observability} problem, rather than a conventional missing-modality problem, and formalize a privileged-information setting in which RNA is available only during training.
\item We propose MoPE, a pathway-structured privileged-distillation framework that organizes RNA supervision through the Hallmark~50 pathways and transfers it to pathway-indexed histology experts via \textit{memory-usage distillation} over a shared latent memory basis, avoiding direct RNA--WSI feature identity.
\item We demonstrate consistent WSI-only improvements across five biomarker classification and five survival prediction tasks on TCGA, and validate generalization on two independent external breast cancer cohorts.
\item We provide pathway-usage analyses and a human-audited visual inspection workflow that make model behaviour inspectable in a biologically organized vocabulary, while supporting pathway-indexed readouts.
\end{itemize}

\section{Related Work}\label{sec:related}

\subsection{Histology-only WSI modelling}

Weakly supervised whole-slide image (WSI) modelling is commonly formulated as a multiple-instance learning problem, in which a slide is represented as a bag of patch-level features. Attention-based MIL \citep{ilse2018attention} and transformer-style MIL \citep{shao2021transmil} models have enabled slide-level prediction by learning to aggregate heterogeneous patch evidence into a patient-level representation. More recent WSI models further improve aggregation through prototype learning \citep{guo2025bpmambamil, koyun2025h}, modern mamba architecture \citep{CHENG2026113442}, or mixture-of-experts transformations \citep{shao2026mixture}. These models satisfy the requirement for H\&E-only deployment. However, training from histology alone may leave valuable transcriptomic information unused. MoPE retains the deployment advantage of WSI-only inference while using paired RNA profiles during training to provide pathway-structured molecular supervision for the histology branch.

\subsection{Pathology--omics information fusion}

Multimodal computational pathology has explored the fusion of histology with transcriptomic data for cancer diagnosis, prognosis, and survival prediction. Representative pathology--omics fusion models combine WSI features with molecular features through tensor fusion \citep{chen2020pathomic}, co-attention \citep{Chen_2021_ICCV}, optimal transport, prototype-based fusion \citep{pmlr-v235-song24b}, or pathway-aware multimodal transformers \citep{yan2025pathway}. These methods show that histology and omics provide complementary information, especially for endpoints involving tumour biology and patient outcome. A key limitation for routine clinical deployment is that many multimodal methods rely on the paired input, and this requirement can hinder deployment in real clinical workflows, where molecular profiling may be unavailable. In contrast, MoPE treats RNA as a privileged training modality: paired WSI--RNA data are used during model development, but the deployed model receives only H\&E WSIs.

\subsection{Privileged learning and distillation with missing molecular modalities}

Knowledge distillation provides a natural framework for using information from a stronger or richer teacher model while preserving a simpler student model at inference \citep{hinton2015distilling}. In computational pathology, recent studies have used multimodal teachers to improve WSI-only students through adaptive teaching \citep{xing2024comprehensive}, histo-genomic association \citep{10830530}, multimodal knowledge decomposition \citep{zhang2026disentangled}, or disentangled learning \citep{zhang2025multi}. These approaches address the practical problem that omics may be unavailable in routine clinical scenarios. However, many distillation objectives rely on direct feature matching between histopathology and omics data. Such objectives can implicitly assume that molecular representations are fully transferable to histology. To address this partial observability problem, MoPE uses an indirect memory-usage distillation objective, aligning RNA pathway tokens and WSI pathway experts in the latent space rather than forcing their embeddings to become identical. This paper is an extension version of our last work \citep{guo2026momentum}, and in this version, we make the model more comprehensive to be applied in multiple clinical scenarios.

\section{Method}\label{sec:method}

\subsection{Problem setting and overview}

\begin{figure}[htbp]
\prfloatbody
\centering
\includegraphics[width=\textwidth]{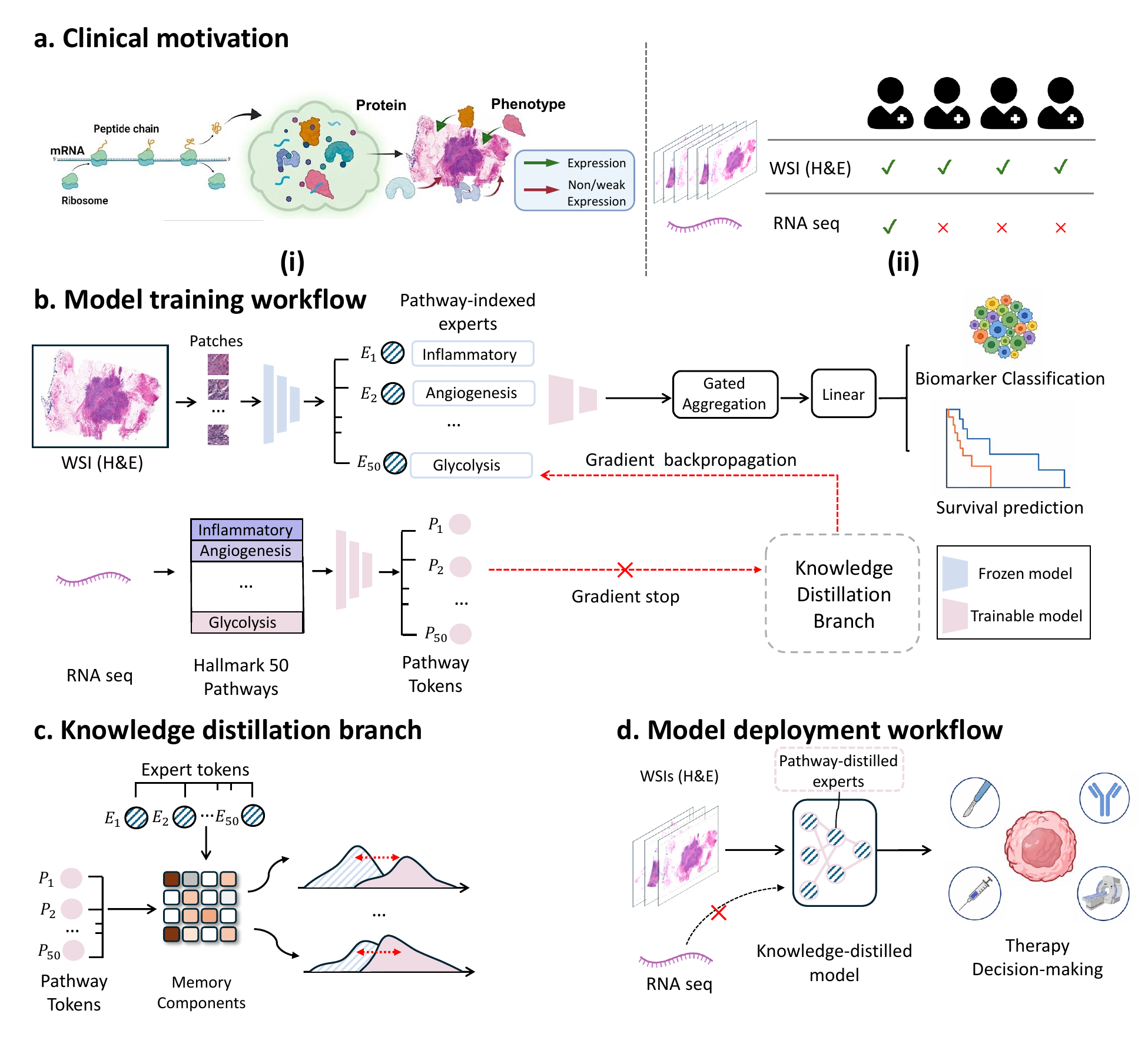}
\caption{
\textbf{Overview of MoPE for pathway-structured privileged distillation.}
\textbf{a}, Clinical motivation. RNA expression, protein abundance and tissue phenotype are connected through a complex and non-linear biological cascade, producing a many-to-many relation between molecular measurements and H\&E morphology. This presents the partial observability: H\&E morphology captures only the tissue-level consequences of some RNA-associated programmes and cannot be assumed to recover the full molecular state of a tumour. RNA profiles can refine cancer phenotyping and risk assessment, but molecular measurements are not consistently available in routine care.
\textbf{b}, Model training workflow. H\&E patches are processed by pathway-indexed experts, whereas RNA expression profiles are organized into Hallmark pathway tokens that provide privileged supervision during training.
\textbf{c}, Knowledge-distillation branch. Matched WSI expert tokens and RNA pathway tokens are encouraged to use a shared memory basis in similar ways, providing a soft distillation objective rather than direct feature imitation.
\textbf{d}, Model deployment workflow. After training, the RNA branch is removed and the pathway-distilled histology branch performs downstream prediction from WSIs alone.
}
\label{fig:model_overview}
\end{figure}

This study addresses a privileged-information setting in which paired whole-slide image (WSI) and bulk RNA sequence data are available during training, whereas only WSIs are available during deployment. The RNA branch is therefore used to provide pathway-structured supervision during model development and is removed at inference. The model performs all downstream predictions from H\&E WSI features alone. The overall framework is shown in Fig.~\ref{fig:model_overview}.

\subsection{WSI branch: pathway-indexed morphology tokens}

Let
$\mathbf X \in \mathbb R^{B \times N \times D_{wsi}}$
denote WSI patch features extracted by a frozen pathology foundation model \citep{chen2024towards}, where $B$ is the batch size, $N$ is the number of patches, and $D_{wsi}$ is the patch feature dimension. We used the Hallmark~50 pathways as the RNA-branch vocabulary, so the WSI branch contains $K=50$ pathway-indexed experts, denoted by $E_1,\ldots,E_K$. Each expert corresponds to one Hallmark pathway and contains $L$ learnable morphology slots. The slots allow one pathway-indexed expert to collect multiple candidate visual patterns before producing a pathway-level WSI token.
For pathway expert $k$, we define a set of learnable slots $\mathbf S_k = [\mathbf s_{k,1}, \ldots, \mathbf s_{k,L}]^\top \in \mathbb R^{L \times D_{\mathrm{wsi}}}$.
The expert computes slot-to-patch routing weights by comparing each learnable slot with linearly projected patch features:
\begin{equation}
\mathbf A_k
=
\operatorname{softmax}_{N}
\left(
\mathbf S_k(\mathbf X\mathbf W_{\mathrm{key}})^{\top}
\right),
\qquad
\mathbf A_k \in \mathbb R^{L \times N}.
\end{equation}
where $\mathbf W_{\mathrm{key}}\in\mathbb R^{D_{\mathrm{wsi}}\times D_{\mathrm{wsi}}}$ is a learnable key projection and the softmax is applied over the patch dimension.
Each row of $\mathbf A_k$ is the attention distribution of one morphology slot over all patches. The corresponding slot embeddings are
\begin{equation}
\mathbf H_k
=
\mathbf A_k\mathbf X,
\qquad
\mathbf H_k
=
[\mathbf h_{k,1}, \ldots, \mathbf h_{k,L}]^\top
\in \mathbb R^{L \times D_{\mathrm{wsi}}}.
\end{equation}
Thus, $\mathbf h_{k,\ell}$ summarizes the WSI evidence selected by slot $\ell$ in pathway expert $k$.

\subsection{Low-rank expert transformation and slide aggregation}

After slot-level morphology extraction, each pathway-indexed expert contains $L$ slot embeddings, where each slot summarizes one candidate morphology pattern from the WSI. Assigning an independent full-rank projection to every expert would overfit limited paired WSI-RNA cohorts. Inspired by low-rank adaptation \citep{hu2022lora}, we use a shared-to-specialized low-rank expert transform. 

For slot $\ell$ in expert $k$, we compute $\mathbf{z}_{k,\ell} = \rho \left( \mathbf{h}_{k,\ell} \mathbf{W}_D \mathbf{U}_k \right)$, where $\mathbf h_{k,\ell}$ is the slot embedding, $\mathbf W_D \in \mathbb R^{D_{wsi} \times r}$ is a shared down-projection, $\mathbf U_k \in \mathbb R^{r \times D_{shared}}$ is the expert-specific projection, $r \ll D_{wsi}$, and $\rho(\cdot)$ denotes layer normalization. This parameterization is intended to encourage reuse of low-rank morphology features across pathways, such as tumour cellularity, stromal organization, necrosis, immune infiltration, or glandular architecture. The expert-specific matrix $\mathbf U_k$ then adapts the shared projection to the semantic context of pathway $k$. 

The transformed slots are then combined within each expert by a learnable slot gate.
For each transformed slot $\mathbf z_{k,\ell}$, we compute a scalar gate logit using a two-layer MLP: $u_{k,\ell}
=
\mathbf w_s^\top
\left(
\operatorname{ReLU}
\left(
\mathbf W_s \mathbf z_{k,\ell}
\right)
\right)$,
where $\mathbf W_s \in \mathbb R^{D_{shared}\times D_{shared}}$ and
$\mathbf w_s \in \mathbb R^{D_{shared}}$ are shared across experts and slots.
The slot weights are normalized within each pathway expert:
\begin{equation}
\beta_{k,\ell}
=
\frac{\exp(u_{k,\ell})}
{\sum_{\ell'=1}^{L}\exp(u_{k,\ell'})},
\qquad
\sum_{\ell=1}^{L}\beta_{k,\ell}=1.
\end{equation}
The pathway-level WSI token is then obtained as
$\mathbf e_k^{wsi}
=
\sum_{\ell=1}^{L}
\beta_{k,\ell}\mathbf z_{k,\ell}$. The resulting token $\mathbf e_k^{wsi} \in \mathbb R^{D_{shared}}$ summarizes WSI morphology evidence for pathway $k$.
Here, $\beta_{k,\ell}$ measures how much the $\ell$-th morphology slot contributes to pathway expert $k$ for the current patient. 

After obtaining the expert token for each pathway, we aggregate these pathway-level tokens into a slide representation:
\begin{equation}
\mathbf z_{slide}
=
\sum_{k=1}^{K}
\alpha_k \mathbf e_k^{wsi},
\qquad
\sum_{k=1}^{K}\alpha_k=1.
\end{equation}
The expert weights $\alpha_k$ are learned from a two-layer
MLP and reflect the relative contribution of different pathway-aligned morphology summaries to the final prediction. The slide representation is then passed to a linear prediction head for the downstream task.

Overall, this module performs a two-stage compression of WSI morphology. The slot gate first selects relevant morphology patterns within each pathway expert, and the expert gate then selects relevant pathway-level summaries at the slide level. This delayed aggregation avoids prematurely collapsing heterogeneous patch evidence and gives the model an inspectable intermediate representation at the pathway level.

\subsection{RNA branch: encoding molecular information}

The RNA branch is used only during training to provide pathway-level privileged targets. We first organize each RNA profile into Hallmark pathway groups, $\mathbf R \in \mathbb R^{B \times K \times D_{rna}}$, where $\mathbf R_{b,k}$ denotes the expression vector associated with pathway $k$ for patient $b$. Each pathway vector is projected into the shared latent space:
\begin{equation}
\mathbf e^{rna}_{b,k}
=
f_{rna}(\mathbf R_{b,k}),
\qquad
\mathbf e^{rna}_{b,k}\in\mathbb R^{D_{shared}}.
\end{equation}
In our implementation, $f_{rna}$ is a linear projection followed by layer normalization. These unmasked RNA pathway tokens are used as the teacher targets for memory-usage distillation.

We also include a lightweight masked reconstruction auxiliary loss on the RNA branch. During training, a subset of pathway vectors is replaced by zeros before projection: $\tilde{\mathbf R}_{b,k}
=
(1-\Omega_{b,k})\mathbf R_{b,k}$,
where $\Omega_{b,k}=1$ indicates a masked pathway. The masked RNA tokens are decoded back to the pathway-gene space, $\hat{\mathbf R}_{b,k}
=
g_{dec}(f_{rna}(\tilde{\mathbf R}_{b,k}))$.

The reconstruction loss is computed only on masked pathways:
\begin{equation}
\mathcal L_{rec}
=
\frac{1}{|\Omega|D_{rna}}
\sum_{b=1}^{B}
\sum_{k=1}^{K}
\Omega_{b,k}
\left\|
\hat{\mathbf R}_{b,k}
-
\mathbf R_{b,k}
\right\|_2^2.
\end{equation}
This auxiliary term regularizes the RNA projection and decoder, whereas the distillation targets are computed from the original unmasked RNA pathway vectors.

\subsection{Distillation in the latent space}

Directly matching WSI and RNA representations imposes a strong assumption and it may hinder model training because the student is penalized for failing to recover modality-private RNA signals. We therefore use a memory bank as a shared latent reference system for a soft distillation: $\mathbf C \in \mathbb R^{M \times D_{shared}}$, where each row $\mathbf c_m$ is a latent basis vector.

For each matched pair of WSI expert token $\mathbf e_k^{wsi}$ and RNA pathway token $\mathbf e_k^{rna}$, we compute their usage/similarity distributions over the shared memory bases: $\mathbf {prob}_k^{wsi}
=
\operatorname{softmax}
\left(
\mathbf e_k^{wsi}\mathbf C^{\top}
\right)$,
$
\mathbf {prob}_k^{rna}
=
\operatorname{softmax}
\left(
\mathrm{sg}(\mathbf e_k^{rna})\mathbf C^{\top}
\right)$, where $\mathrm{sg}(\cdot)$ denotes the stop-gradient operation. The stop-gradient prevents the RNA teacher target from being updated by the student-side task loss or by the distillation objective. As a result, the RNA branch provides a stable pathway-level target.

The distillation loss is defined as
$
\mathcal L_{KD}
=
\frac{1}{K}
\sum_{k=1}^{K}
D_{KL}
\left(
\mathbf {prob}_k^{rna}
\;\|\;
\mathbf {prob}_k^{wsi}
\right).$ 
This objective asks the WSI expert token and the RNA pathway token to use the shared latent bases similarly, rather than requiring their embeddings to coincide. This memory-usage distillation is intended to transfer RNA-side structure through shared latent-basis usage while allowing each modality to retain private information.

To stabilize the shared memory basis, we use a vector-quantization style memory regularization \citep{gray1984vector}. For each WSI expert token, we assign it to the nearest memory basis. Specifically, after flattening all WSI expert tokens in a mini-batch into
$\{\mathbf e_i^{wsi}\}_{i=1}^{I}$, where $I=BK$, the nearest memory index is
$q_i
=
\arg\min_{m\in\{1,\ldots,M\}}
\left\|
\mathbf e_i^{wsi}-\mathbf c_m
\right\|_2^2$.
The memory regularization contains a commitment term and an orthogonality term:
\begin{equation}
\mathcal L_{\mathrm{mem}}
=
\frac{1}{I}
\sum_{i=1}^{I}
\left\|
\mathbf e_i^{wsi}
-
\mathbf c_{q_i}
\right\|_2^2
+
\sum_{m=1}^{M}
\sum_{m'\ne m}^{M}
\left(
\frac{
\mathbf c_m^\top \mathbf c_{m'}
}{
\|\mathbf c_m\|_2 \|\mathbf c_{m'}\|_2
}
\right)^2 .
\end{equation}
The first term encourages WSI expert tokens to remain close to their assigned memory bases. The second term discourages different memory bases from becoming redundant by penalizing pairwise cosine similarity among codebook entries.

Before optimization, we initialize the memory by applying k-means on pooled WSI embeddings from the training set, and use the resulting $M$ cluster centers to initialize the memory base.

\subsection{Training objective and inference}

The final objective combines task supervision, memory-usage distillation, masked pathway reconstruction, memory regularization, and slot diversity:
\begin{equation}
\mathcal L
=
\lambda_{task}\mathcal L_{task}
+
\lambda_{KD}\mathcal L_{KD}
+
\lambda_{rec}\mathcal L_{rec}
+
\lambda_{mem}\mathcal L_{mem}
+
\lambda_{div}\mathcal L_{div}.
\end{equation}

The task loss $\mathcal L_{task}$ is cross-entropy for classification. For survival prediction, we used a negative log-likelihood (NLL) loss.

Further, we regularize the slot attention:
$
\mathcal L_{div}
=
\frac{1}{K}
\sum_{k=1}^{K}
\frac{1}{L(L-1)}
\sum_{\ell \ne \ell'}
\operatorname{cos}
\left(
\mathbf A_{k,\ell},
\mathbf A_{k,\ell'}
\right)$, where $\mathbf A_{k,\ell}$ is the slot-to-patch attention vector of slot $\ell$ in expert $k$.

In our default setting, the loss weights are set as: $\lambda_{task}=1.0$, $\lambda_{mem}=0.05$, $\lambda_{rec}=0.01$, and $\lambda_{div}=0.1$. Specifically, for classification tasks, $\lambda_{KD}$ is $1.0$, while for survival prediction, $\lambda_{KD}$ is $0.5$, with all other parameters remaining consistent.

\section{Experiments}\label{sec:experiments}

\subsection{Datasets}\label{sec:datasets}

We evaluated MoPE using The Cancer Genome Atlas (TCGA) \url{https://portal.gdc.cancer.gov}, encompassing breast (TCGA-BRCA, $n = 1{,}023$), lung (TCGA-LUAD, $n = 510$), glioblastoma/lower-grade glioma (TCGA-GBMLGG, $n = 878$), stomach (TCGA-STAD, $n = 363$), and kidney (TCGA-KIRC, $n = 498$) cancers.

For clinical classification, the endpoint including Oncotype DX (ODX) risk and Progesterone Receptor (PR) status in breast cancer, Hypoxia-Inducible Factor and Epidermal Growth Factor Receptor (EGFR) status in lung cancer, and mesenchymal state in glioma. Survival prediction was evaluated across five cancer types to test cross-tissue performance. Matched RNA profiles were retrieved from the UCSC Xena platform \citep{goldman2020visualizing}, and the 50 Hallmark gene sets obtained from the Molecular Signatures Database (MSigDB) \citep{subramanian2005gene}. Detailed statistics for all TCGA tasks are summarized in Supplementary Table S1, with additional clinical context provided in Supplementary Note~1.

To test the generalizability of the model, we used two independent external cohorts for ODX risk prediction ($n = 1{,}645$). These comprised 1,123 whole-slide images (WSIs) of breast cancer from The Ohio State University Wexner Medical Center, and 522 WSIs from the Dartmouth Breast Cancer Recurrence Risk Dataset \citep{goyal2024multi}. These geographically and institutionally distinct cohorts provide comprehensive evaluation of the proposed method under the real clinical scenario.

\subsection{Comparison methods}

We compared our method with four groups of baselines.

First, for the omics-only setting, we used the Self-Normalizing Network (SNN) \citep{klambauer2017self} and GPNet \citep{lu2025classification} as the RNA-based methods. These models use omics features only and do not use WSI information.

Second, for the histopathology-only setting, we included representative multiple instance learning methods, including attention based MIL (AttMIL) \citep{ilse2018attention}, DSMIL \citep{li2021dsmil}, DTFD-MIL \citep{li2021dtfdmil}, TransMIL \citep{shaoTransMILTransformerBased2021}, and WIKG \citep{liDynamicGraphRepresentation2024}. Further, we considered MAMMOTH \citep{shao2026mixture}, a mixture-of-experts based method in computational pathology. Following the official implementation, we integrated MAMMOTH with AttMIL and TransMIL, denoted as AttMIL-MoE and TransMIL-MoE in the following sections.

Third, for knowledge distillation methods, we compared TDC \citep{xing2024comprehensive}, G-HANet \citep{10830530}, MKD \citep{zhangMultimodalKnowledgeDecomposition2025}, and DMML \citep{zhang2026disentangled}. These methods share the same scope as our work: both WSI and RNA are available during training, while remaining RNA-free inference.

Finally, we included multimodal methods that use paired WSI and omics data during both training and inference as a performance upper bound, including DMML$_t$ \citep{zhang2026disentangled}, MMP \citep{pmlr-v235-song24b}, LD-CVAE \citep{zhou2025robustmultimodalsurvivalprediction} and PAMT \citep{11172339}. Here, DMML$_t$ denotes the multimodal teacher model from the DMML paper.

\subsection{Implementation details}

Following established protocols \citep{zhang2025standardizing}, we segmented the tissue foreground of each WSI and extracted $896 \times 896$ pixel patches from tissue foreground regions. For feature extraction, we used the frozen foundation model UNI v2 \citep{chen2024towards}, yielding patch-level representations of dimension $D_{wsi}=1536$.

For the pathway-level RNA representation, we used the 50 Hallmark gene sets and retained genes that overlapped with the original RNA expression profile ($D=20{,}531$), resulting in 4,188 unique Hallmark-associated genes. For each patient, raw expression values were first transformed by $\log_2(x+1)$ and then standardized gene-wise across primary tumor samples to obtain z-scores. We then constructed a $50 \times 4188$ pathway-gene matrix. Entries for genes belonging to the corresponding pathway were filled with their expression values, whereas non-member genes were zero-masked. 

All models were trained with AdamW (learning rate $2\times10^{-4}$, weight decay $1\times10^{-4}$, batch size 1). For classification, we used patient-level stratified 5-fold cross-validation with a $6{:}2{:}2$ train/validation/test split and trained for 30 epochs. For external evaluation, TCGA-BRCA was split $9{:}1$ into training and validation sets in each fold, and the trained weight was directly evaluated on two independent cohorts without fine-tuning. Survival models followed the same 5-fold protocol and were trained for 20 epochs. Results are reported as mean $\pm$ standard deviation across five folds. All experiments used a single NVIDIA H100 GPU.

\subsection{Human-audited visual inspection}

For the visual inspection analysis, we selected representative correct and failed BRCA-ODX predictions and traced them through expert-level attention maps, slot-level attention maps and high-attention H\&E patch tiles. A large language model (\textit{ChatGPT 5.5, utilized in May 2026}) was used only to convert these model-derived visual materials into a structured morphology summary using the prompt template in Supplementary Note~3. A clinical expert then reviewed both the summary and the underlying image evidence using predefined criteria for morphological accuracy, image support, pathway compatibility and ambiguity. The analysis was designed to assess whether pathway-indexed model readouts were visually plausible and to identify failure modes.

\section{Results}\label{sec:results}

We evaluated MoPE across five biomarker classification tasks and five survival prediction tasks from TCGA-BRCA, TCGA-LUAD, TCGA-GBMLGG, TCGA-STAD and TCGA-KIRC, and then tested breast cancer Oncotype DX (ODX) prediction in two independent external cohorts, OSUWMC ($n = 1{,}123$) and Dartmouth ($n = 522$)  (Fig.~\ref{fig:internal_validation} and Fig.~\ref{fig:external_validation}).

\subsection{Pathway-structured distillation improves WSI-only prediction}

\begin{figure}[htbp]
\prfloatbody
\centering
\includegraphics[width=\textwidth, height=0.78\textheight, keepaspectratio]{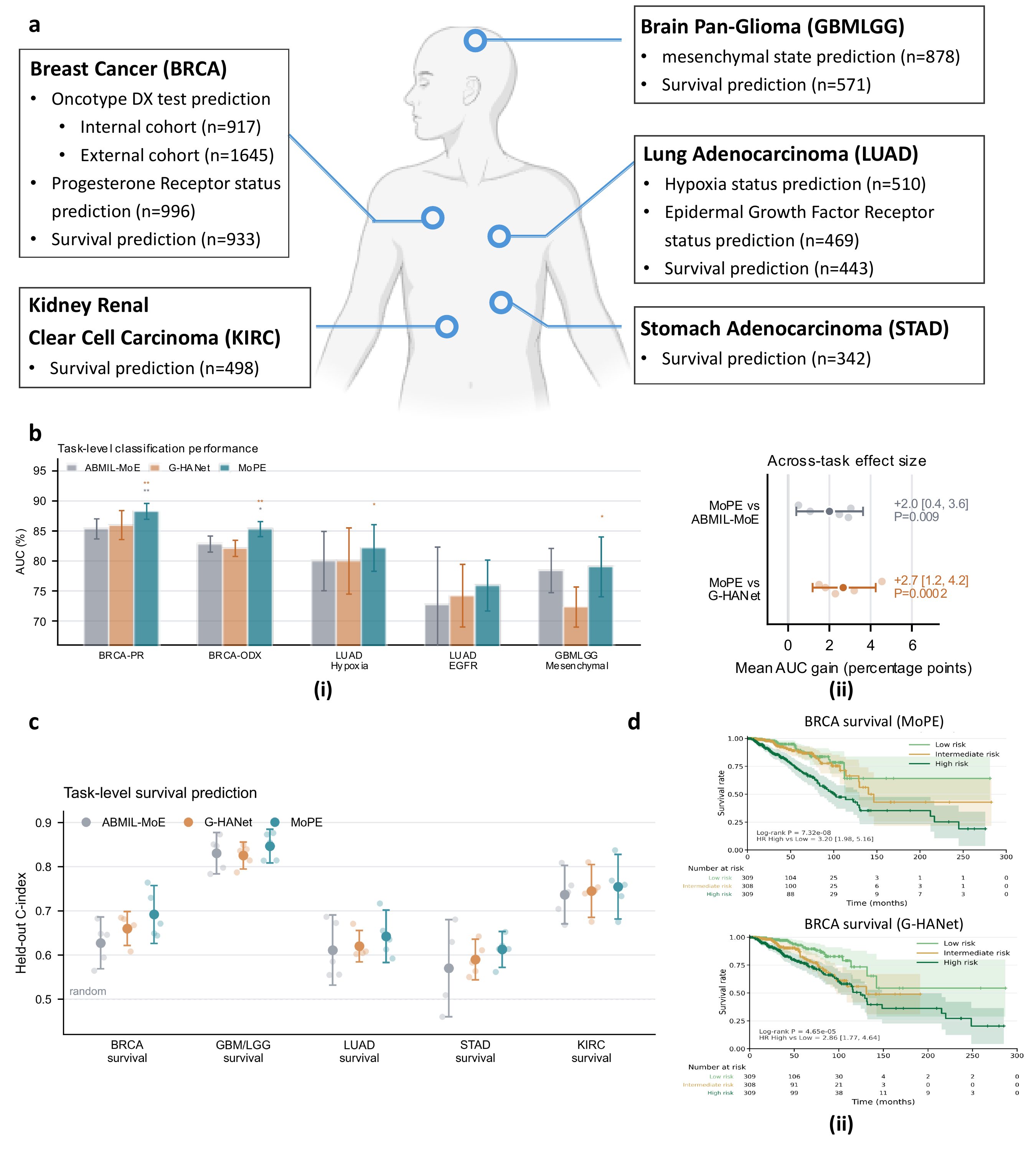}
\caption{
\textbf{Internal evaluation across biomarker classification and survival prediction tasks.}
\textbf{a}, Overview of the internal TCGA tasks across breast, brain, lung, stomach and kidney cancer cohorts.
\textbf{b}, Biomarker classification across five endpoints. (i) AUC (\%) for AttMIL-MoE, G-HANet and MoPE; bars show five-fold means and error bars s.d. Asterisks indicate nominal patient-paired bootstrap comparisons with MoPE. (ii) Across-task AUC gains of MoPE over each baseline; large points show mean gains, bars show 95\% paired bootstrap confidence intervals, and faint points indicate task-level gains.
\textbf{c}, Survival prediction across five TCGA cohorts. Small points show individual folds, large points fold means, and error bars 95\% bootstrap confidence intervals.
\textbf{d}, Kaplan--Meier overall-survival curves by predicted-risk tertiles for MoPE and G-HANet in TCGA-BRCA. Shading shows 95\% confidence intervals; HRs were estimated by Cox regression and risk strata compared by log-rank tests.
}
\label{fig:internal_validation}
\end{figure}

We first evaluated whether transcriptomic supervision improves prediction when only H\&E WSIs are available at inference. Across five biomarker classification tasks, MoPE achieved the highest average WSI-only AUC ($82.36\%$), compared with $79.84\%$ for AttMIL-MoE and $79.03\%$ for G-HANet (Table~\ref{tab:main_classification_merged}). Across-task analysis showed mean AUC gains of $2.0$ percentage points over AttMIL-MoE (95\% CI, $0.4$--$3.6$; $P=0.009$) and $2.7$ points over G-HANet (95\% CI, $1.2$--$4.2$; $P=0.0002$) (Fig.~\ref{fig:internal_validation}b(ii)). Because RNA was used only during training, these gains reflect improved WSI-only representations rather than test-time multimodal input.

MoPE also improved survival prediction across five TCGA cohorts, achieving an average C-index of $70.96\%$ versus $67.48\%$ and $68.81\%$ for the strongest histology-only and distillation baselines, respectively (Fig.~\ref{fig:internal_validation}c; Table~\ref{tab:survival}). In TCGA-BRCA, predicted-risk tertiles showed stronger survival separation with MoPE (log-rank $P=7.32\times10^{-8}$; HR $=3.2$, 95\% CI $1.98$--$5.16$) than with G-HANet ($P=4.65\times10^{-5}$; HR $=2.86$, 95\% CI $1.77$--$4.64$) (Fig.~\ref{fig:internal_validation}d). Additional Kaplan--Meier analyses across survival cohorts are provided in Supplementary Fig.~S2.

Finally, MoPE approached multimodal methods that require RNA at inference on several endpoints, including BRCA-PR and LUAD-EGFR (Table~\ref{tab:main_classification_merged}), suggesting that pathway-structured distillation recovers part of the multimodal performance benefit while retaining WSI-only deployment.

\begingroup
\prtablebody
\setlength{\LTleft}{\fill}
\setlength{\LTright}{\fill}
\setlength{\tabcolsep}{2.5pt}
\renewcommand{\arraystretch}{1.0}
\begin{longtable}{lcccccc}
\caption{\textbf{Multi-dataset and external clinical classification AUC (\%).} 
Values are mean $\pm$ s.d. across five folds. Methods are grouped by inference modality. 
Ext-ODX denotes the pooled OSUWMC and Dartmouth external cohorts. 
Gene-only and multimodal methods are not applicable to Ext-ODX because paired omics data are unavailable. 
Bold and underline denote the best and second-best performance within each group, respectively.}
\label{tab:main_classification_merged}\\

\toprule
\textbf{Method} &
\makecell{PR} &
\makecell{ODX} &
\makecell{Hypoxia} &
\makecell{EGFR} &
\makecell{Mesenchymal} &
\makecell{Ext-ODX} \\
\midrule
\endfirsthead
\multicolumn{7}{c}{\textit{Table \thetable{} continued}}\\
\toprule
\textbf{Method} &
\makecell{PR} &
\makecell{ODX} &
\makecell{Hypoxia} &
\makecell{EGFR} &
\makecell{Mesenchymal} &
\makecell{Ext-ODX} \\
\midrule
\endhead
\midrule
\multicolumn{7}{r}{\textit{Continued on next page}}\\
\endfoot
\bottomrule
\endlastfoot
\multicolumn{7}{c}{\textbf{Gene-only models}} \\

SNN &
\rsu{90.59}{0.64} &
\rsb{91.88}{0.75} &
\rsu{90.66}{0.55} &
\rsb{85.87}{7.23} &
\rsu{92.19}{3.53} &
-- \\

GPNet &
\rsb{91.74}{0.23} &
\rsu{90.70}{1.57} &
\rsb{90.71}{1.53} &
\rsu{85.21}{5.34} &
\rsb{93.22}{4.10} &
-- \\

\midrule
\multicolumn{7}{c}{\textbf{Histology-only models}} \\

AttMIL &
\rs{84.52}{2.31} &
\rs{79.31}{2.54} &
\rs{79.02}{4.10} &
\rs{70.22}{5.71} &
\rs{73.71}{2.10} &
\rs{75.42}{1.73} \\

DSMIL &
\rs{81.64}{2.83} &
\rs{78.70}{2.09} &
\rs{78.81}{3.22} &
\rs{70.01}{10.94} &
\rs{74.25}{2.24} &
\rs{75.31}{3.06} \\

DTFD-MIL &
\rsu{85.25}{1.02} &
\rs{77.91}{3.55} &
\rs{78.24}{3.55} &
\rs{70.56}{7.30} &
\rs{76.47}{3.41} &
\rsu{76.01}{2.89} \\

TransMIL &
\rs{83.90}{2.40} &
\rs{79.10}{2.17} &
\rs{76.36}{4.39} &
\rs{64.83}{11.10} &
\rs{67.54}{4.87} &
\rs{71.79}{2.44} \\

WIKG &
\rs{84.90}{3.04} &
\rs{78.36}{3.74} &
\rs{78.61}{5.57} &
\rsu{72.01}{4.86} &
\rs{73.51}{4.94} &
\rs{73.39}{5.41} \\

AttMIL-MoE &
\rsb{85.34}{1.50} &
\rsb{82.80}{1.20} &
\rsb{79.97}{4.42} &
\rsb{72.68}{8.60} &
\rsb{78.40}{3.29} &
\rs{75.28}{1.77} \\

TransMIL-MoE &
\rs{85.10}{2.14} &
\rsu{79.84}{1.51} &
\rsu{79.02}{4.06} &
\rs{70.89}{9.21} &
\rsu{77.60}{3.15} &
\rsb{76.30}{3.29} \\

\midrule
\multicolumn{7}{c}{\makecell{\textbf{Knowledge distillation models}}} \\

TDC &
\rs{84.70}{5.36} &
\rs{81.01}{2.23} &
\rs{78.18}{3.33} &
\rs{72.27}{8.81} &
\rs{75.54}{3.58} &
\rs{75.56}{3.12} \\

G-HANet &
\rsu{85.97}{2.15} &
\rsu{82.10}{1.35} &
\rsu{79.62}{4.29} &
\rsu{74.21}{5.22} &
\rs{73.23}{2.65} &
\rsu{77.30}{1.92} \\

MKD &
\rs{85.00}{2.35} &
\rs{80.14}{1.55} &
\rs{79.01}{5.11} &
\rs{74.01}{6.66} &
\rsu{77.94}{4.49} &
\rs{76.56}{3.99} \\

DMML &
\rs{82.50}{2.34} &
\rs{81.57}{3.43} &
\rs{79.11}{5.21} &
\rs{72.91}{7.73} &
\rs{76.28}{3.22} &
\rs{76.14}{2.31} \\

Ours &
\rsb{88.27}{1.32} &
\rsb{85.38}{1.22} &
\rsb{82.17}{3.88} &
\rsb{75.90}{4.25} &
\rsb{80.09}{5.21} &
\rsb{79.88}{3.63} \\

\midrule
\multicolumn{7}{c}{\textbf{Multimodal models}} \\

DMML$_t$ &
\rsu{87.84}{0.61} &
\rsu{88.22}{4.13} &
\rs{93.34}{6.90} &
\rsb{78.51}{5.51} &
\rsb{91.75}{1.56} &
-- \\

MMP &
\rs{87.00}{3.32} &
\rs{87.30}{5.51} &
\rs{92.21}{6.51} &
\rs{76.24}{4.38} &
\rs{89.24}{3.32} &
-- \\

LD-CVAE &
\rsb{88.54}{3.41} &
\rs{87.13}{3.15} &
\rsb{94.25}{1.42} &
\rsu{78.30}{4.79} &
\rs{88.44}{4.16} &
-- \\

PAMT &
\rs{85.17}{4.82} &
\rsb{88.28}{3.44} &
\rsu{93.84}{1.78} &
\rs{75.40}{9.29} &
\rsu{90.01}{4.32} &
-- \\

\end{longtable}
\endgroup

\begingroup
\prtablebody
\setlength{\LTleft}{\fill}
\setlength{\LTright}{\fill}
\setlength{\tabcolsep}{2.5pt}
\renewcommand{\arraystretch}{1.0}
\begin{longtable}{lccccc}
\caption{\textbf{Survival prediction performance on five TCGA cohorts measured by C-index (\%).} Values are mean $\pm$ s.d. across five folds. Methods are grouped by inference modality. Bold and underline denote the best and second-best performance within each group, respectively.}
\label{tab:survival}\\
\toprule
\textbf{Method} &
\makecell{BRCA} &
\makecell{LUAD} &
\makecell{STAD} &
\makecell{GBMLGG} &
\makecell{KIRC} \\
\midrule
\endfirsthead
\multicolumn{6}{c}{\textit{Table \thetable{} continued}}\\
\toprule
\textbf{Method} &
\makecell{BRCA} &
\makecell{LUAD} &
\makecell{STAD} &
\makecell{GBMLGG} &
\makecell{KIRC} \\
\midrule
\endhead
\midrule
\multicolumn{6}{r}{\textit{Continued on next page}}\\
\endfoot
\bottomrule
\endlastfoot
\multicolumn{6}{c}{\textbf{Gene-only models}} \\
SNN & \rsu{61.02}{5.54} & \rsu{63.04}{7.21} & \rsu{57.88}{8.31} & \rsu{80.93}{3.65} & \rsu{68.87}{8.12} \\
GPNet & \rsb{66.19}{3.60} & \rsb{75.21}{4.22} & \rsb{61.02}{3.57} & \rsb{84.62}{5.92} & \rsb{74.53}{3.26} \\
\midrule
\multicolumn{6}{c}{\textbf{Histology-only models}} \\
AttMIL & \rs{60.27}{4.23} & \rs{57.99}{4.21} & \rs{57.14}{6.28} & \rs{81.16}{3.25} & \rs{69.63}{3.60} \\
DSMIL & \rs{59.14}{3.28} & \rs{57.01}{4.45} & \rsb{58.05}{6.63} & \rs{82.04}{3.52} & \rs{59.72}{4.61} \\
DTFD-MIL & \rs{60.77}{3.42} & \rs{51.60}{6.93} & \rs{54.32}{7.61} & \rs{80.15}{4.33} & \rs{68.66}{6.07} \\
TransMIL & \rs{58.86}{4.82} & \rs{56.08}{5.52} & \rs{53.45}{5.74} & \rs{81.08}{4.51} & \rs{64.98}{1.45} \\
WIKG & \rsu{63.95}{2.14} & \rs{56.11}{6.42} & \rsu{57.48}{6.42} & \rs{82.23}{4.37} & \rs{68.59}{1.95} \\
AttMIL-MoE & \rs{62.74}{4.21} & \rsb{61.01}{5.72} & \rs{57.00}{7.93} & \rsu{83.04}{3.40} & \rsb{73.64}{4.78} \\
TransMIL-MoE & \rsb{64.43}{3.92} & \rsu{59.90}{13.84} & \rs{53.64}{8.91} & \rsb{83.14}{3.32} & \rsu{73.25}{4.43} \\
\midrule
\multicolumn{6}{c}{\makecell{\textbf{Knowledge distillation models}}} \\
TDC & \rs{65.82}{4.54} & \rs{60.03}{6.61} & \rs{57.63}{7.76} & \rs{81.34}{4.92} & \rs{71.56}{6.11} \\
G-HANet & \rsu{66.04}{2.89} & \rsu{62.01}{2.57} & \rs{59.00}{3.31} & \rsu{82.52}{2.21} & \rsu{74.49}{4.80} \\
MKD & \rs{66.01}{3.91} & \rs{60.39}{4.21} & \rs{57.68}{2.31} & \rs{81.60}{1.11} & \rs{73.15}{5.45} \\
DMML & \rs{64.58}{4.73} & \rs{60.22}{5.32} & \rsu{59.93}{6.78} & \rs{82.04}{5.12} & \rs{72.23}{3.12} \\
Ours & \rsb{69.18}{5.30} & \rsb{64.24}{4.78} & \rsb{61.27}{3.26} & \rsb{84.65}{2.75} & \rsb{75.47}{5.89} \\
\midrule
\multicolumn{6}{c}{\textbf{Multimodal models}} \\
DMML$_t$ & \rs{69.86}{6.02} & \rsu{66.06}{5.33} & \rsb{63.14}{3.35} & \rs{85.92}{3.73} & \rs{75.24}{4.41} \\
MMP & \rsb{74.23}{6.35} & \rs{65.02}{6.03} & \rsu{62.38}{4.21} & \rsb{87.04}{4.14} & \rsb{76.15}{5.12} \\
LD-CVAE & \rsu{69.94}{4.31} & \rs{65.47}{4.66} & \rs{62.21}{5.72} & \rsu{86.13}{4.23} & \rs{75.92}{6.15} \\
PAMT & \rs{69.35}{5.97} & \rsb{67.21}{5.98} & \rs{61.39}{3.77} & \rs{85.84}{3.06} & \rsu{76.00}{5.23} \\
\end{longtable}
\endgroup

\subsection{MoPE generalizes to two independent breast cancer patient cohorts}

\begin{figure}[htbp]
\prfloatbody
\centering
\includegraphics[width=\textwidth]{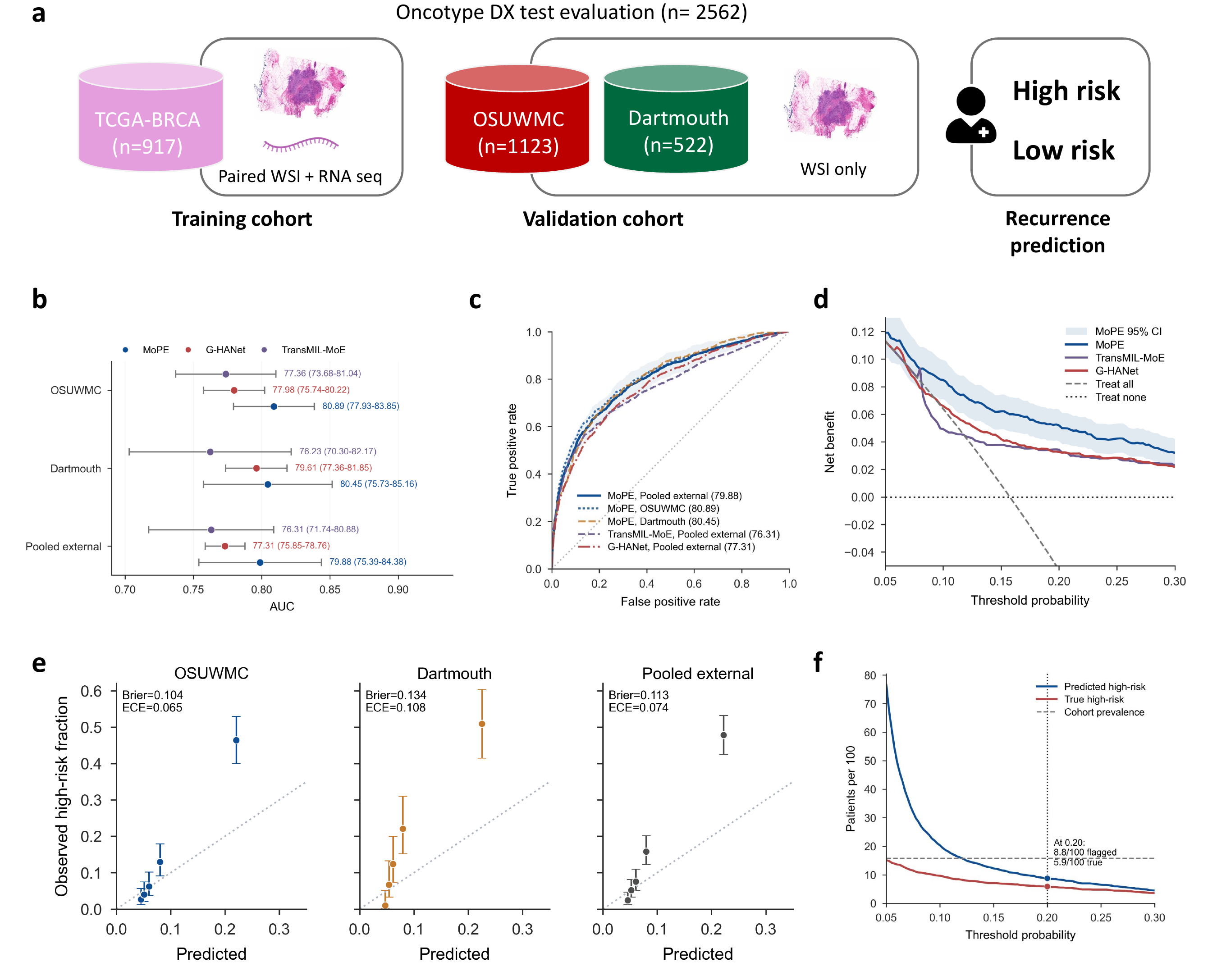}
\caption{
\textbf{External evaluation in two independent patient cohorts.}
\textbf{a,} External evaluation design. MoPE was trained on TCGA-BRCA with paired WSI and RNA and evaluated for Oncotype DX prediction in two independent WSI-only cohorts, OSUWMC ($n=1{,}123$) and Dartmouth ($n=522$).
\textbf{b,} External AUCs for MoPE, G-HANet and TransMIL-MoE. Horizontal bars show 95
\textbf{c,} ROC curves for OSUWMC, Dartmouth and pooled external cohorts. AUC values are shown in parentheses.
\textbf{d,} Decision-curve analysis on the pooled cohort. Curves show net benefit across threshold probabilities; shading indicates the 95
\textbf{e,} Calibration of MoPE after fold-wise Platt scaling. Points show binned predicted versus observed risk with 95
\textbf{f,} Clinical impact curve for MoPE on the pooled cohort, showing predicted and true high-risk patients per 100 patients. The main text reports the operating point at threshold probability 0.20.
}
\label{fig:external_validation}
\end{figure}

We next evaluated MoPE on two independent WSI-only cohorts for Oncotype DX risk prediction: OSUWMC ($n=1{,}123$) and Dartmouth ($n=522$) (Fig.~\ref{fig:external_validation}a). MoPE achieved AUCs of 80.89\% (95\% CI, 77.93--83.85) and 80.45\% (95\% CI, 75.73--85.16), respectively. In the pooled cohort, MoPE reached 79.88\% (95\% CI, 75.39--84.38), outperforming TransMIL-MoE (76.31\%) and G-HANet (77.31\%), with significant paired-bootstrap improvements over both baselines ($P<10^{-3}$; Fig.~\ref{fig:external_validation}b,c).

After Platt scaling fitted only on TCGA validation predictions, MoPE showed monotonic risk calibration across both external sites, with Brier scores of 0.104, 0.134 and 0.113 for OSUWMC, Dartmouth and the pooled cohort, respectively (Fig.~\ref{fig:external_validation}e). Decision-curve analysis further showed higher net benefit than baseline and treat-all/treat-none strategies across the evaluated thresholds (Fig.~\ref{fig:external_validation}d). At a threshold of 0.20, MoPE identified 8.8 patients per 100 as high risk, including 5.9 true high-risk cases (Fig.~\ref{fig:external_validation}f). Together, these results demonstrate robust cross-institutional generalization under WSI-only deployment.

\subsection{Ablations support pathway-structured, indirect distillation}

We next tested whether the performance gains were consistent with the design rationale rather than being explained solely by the addition of molecular supervision or expert capacity. All ablation variants used the same data splits, feature extraction pipeline, task heads and optimization schedule as the full model unless otherwise stated. Replacing memory-usage distillation with direct cosine alignment between matched RNA and WSI pathway embeddings reduced mean classification AUC from $82.36\%$ to $80.03\%$ and mean survival C-index from $70.96\%$ to $68.41\%$ (Table~\ref{tab:ablation_mechanism}). This result is consistent with the partial-observability premise, although it does not by itself prove biological partial observability: direct feature alignment imposes a strong one-to-one correspondence between RNA and WSI representations, whereas the WSI branch observes only an incomplete morphological view of molecular state.

\begin{table}[htbp]
\prfloatbody
\centering
\caption{
\textbf{Ablation of pathway-structured distillation components.}
Results are mean $\pm$ s.d. across five folds, reported as AUC (\%) for classification and C-index (\%) for survival. Direct feature alignment replaces pathway-level memory-usage distillation with cosine alignment between WSI and RNA embeddings.
}
\label{tab:ablation_mechanism}
\setlength{\tabcolsep}{3pt}
\begin{tabular}{lcccc}
\toprule
\textbf{Task}
& \textbf{Full model}
& \makecell{\textbf{Direct feature}\\\textbf{alignment}}
& \makecell{\textbf{w/o Pathway}\\\textbf{sup.}}
& \makecell{\textbf{w/o Slot}\\\textbf{diversity}} \\
\midrule
BRCA-PR      & 88.27$\pm$1.32 & 85.51$\pm$1.74 & 85.91$\pm$1.67 & 87.26$\pm$2.15 \\
BRCA-ODX     & 85.38$\pm$1.22 & 83.15$\pm$1.97 & 80.41$\pm$2.01 & 83.18$\pm$1.97 \\
LUAD-Hypoxia & 82.17$\pm$3.88 & 80.85$\pm$4.96 & 80.11$\pm$5.10 & 81.54$\pm$4.11 \\
LUAD-EGFR    & 75.90$\pm$4.25 & 72.90$\pm$6.25 & 72.80$\pm$8.00 & 73.70$\pm$7.69 \\
GBMLGG-Mes   & 80.09$\pm$5.21 & 77.73$\pm$3.99 & 79.29$\pm$4.29 & 78.02$\pm$4.99 \\
BRCA-Surv    & 69.18$\pm$5.30 & 66.73$\pm$5.47 & 63.30$\pm$3.89 & 64.08$\pm$6.11 \\
LUAD-Surv    & 64.24$\pm$4.78 & 60.39$\pm$4.91 & 59.97$\pm$5.01 & 62.11$\pm$4.57 \\
STAD-Surv    & 61.27$\pm$3.26 & 58.50$\pm$4.92 & 58.02$\pm$4.11 & 60.58$\pm$4.68 \\
GBMLGG-Surv  & 84.65$\pm$2.75 & 83.00$\pm$3.67 & 81.90$\pm$3.31 & 83.19$\pm$3.20 \\
KIRC-Surv    & 75.47$\pm$5.89 & 73.46$\pm$5.91 & 72.21$\pm$4.44 & 74.47$\pm$5.89 \\
\midrule
\textbf{Average} & \textbf{76.66} & 74.22 & 73.39 & 74.81 \\
\bottomrule
\end{tabular}
\end{table}

Removing pathway supervision and slot diversity was also associated with lower performance. Without pathway supervision, mean classification AUC decreased to $79.70\%$ and mean survival C-index to $67.08\%$, consistent with privileged RNA information being more useful when organized through a pathway vocabulary. Without slot diversity, the corresponding averages decreased to $80.74\%$ and $68.89\%$, consistent with multiple morphology slots helping pathway-indexed experts represent heterogeneous tissue evidence rather than collapsing each pathway into a single visual prototype. Supplementary Fig.~S1 further shows that memory and slot capacities affect performance, with overly large configurations reducing average AUC on the evaluated classification tasks.

\subsection{Pathway expert behaviour reflects incomplete RNA--WSI correspondence}

\begin{figure}[htbp]
\prfloatbody
\centering
\includegraphics[width=\textwidth]{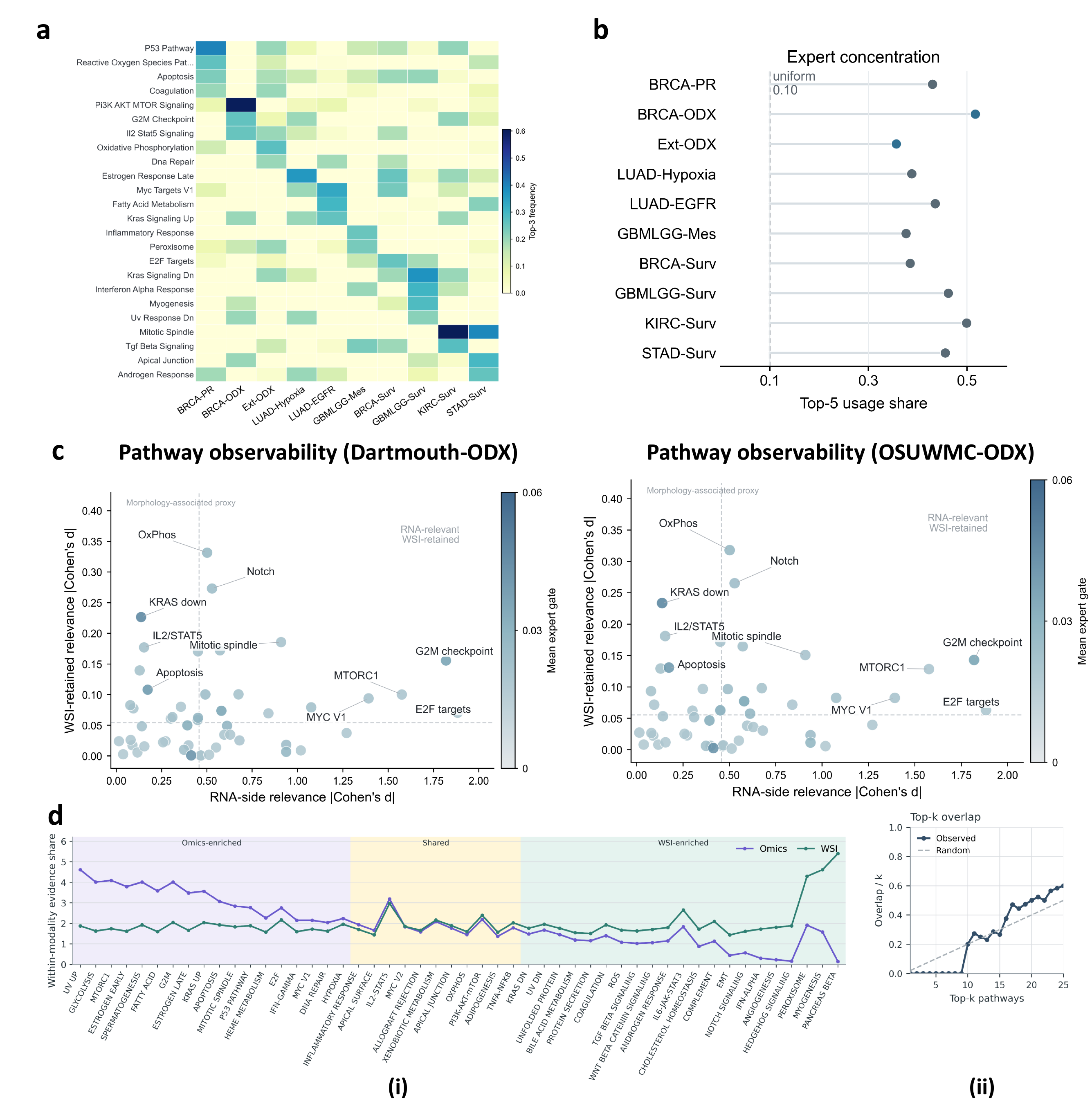}
\caption{
\textbf{Pathway experts provide task-dependent, non-identical WSI and omics readouts.}
\textbf{a}, WSI pathway expert usage across classification, external and survival tasks, shown as top-three selection frequency.
\textbf{b}, Task-wise expert concentration, defined as the fraction of top-three usage captured by the five most frequently selected experts. Dashed line indicates the uniform expectation across 50 pathways.
\textbf{c}, RNA- versus WSI-side pathway relevance in Dartmouth and OSUWMC. Each point is a Hallmark pathway; axes show absolute Cohen's $d$ for RNA pathway scores and WSI gate values, with colour indicating expert frequency. Dashed lines mark pathway-wise medians.
\textbf{d}, Comparison of omics-only GPNet and MoPE pathway profiles for BRCA-ODX. (i) Normalized pathway-score shares. (ii) Top-$k$ ranking overlap; dashed line indicates random expectation ($k/50$).
}
\label{fig:pathway_behavior}
\end{figure}

We next examined whether MoPE used pathway experts uniformly or in a task-dependent manner. Expert-gate patterns differed across tasks (Fig.~\ref{fig:pathway_behavior}a). For each WSI, we identified the three experts with the highest gate values and summarized their selection frequencies at the task level. The five most frequently selected experts accounted for $35.7$--$51.7\%$ of selections across tasks, substantially above the $10\%$ expected under uniform use of 50 Hallmark experts (Fig.~\ref{fig:pathway_behavior}b), indicating concentrated and task-dependent expert usage.

We next compared RNA-level pathway relevance with WSI-side expert usage using the absolute Cohen's $d$ for each Hallmark pathway, calculated from RNA pathway scores and corresponding WSI expert-gate values, respectively (Supplementary Note~2). For the external ODX cohorts, where RNA profiles were unavailable, internal BRCA-ODX RNA relevance served as the reference. RNA- and WSI-derived effect-size patterns showed only partial correspondence across both clinical centers (Fig.~\ref{fig:pathway_behavior}c), suggesting that transcriptomic pathway relevance is not mapped one-to-one onto morphological expert usage.

We further compared MoPE with the omics-only pathway model GPNet~\citep{lu2025classification}. MoPE pathway scores were defined by mean expert-gate values, whereas GPNet scores were derived from pathway-level association statistics after Benjamini--Hochberg correction. Ranking pathways by the difference between normalized WSI and omics scores revealed omics-enriched, shared, and WSI-enriched regimes (Fig.~\ref{fig:pathway_behavior}d(i)). Estrogen-response and metabolic pathways were more prominent in the omics-enriched region, whereas IL2--STAT5 and PI3K--AKT--mTOR-related programmes appeared among shared pathways. WSI-enriched pathways showed stronger histology-derived scores but should not be interpreted as directly measured pathway activity.

Consistently, top-$k$ overlap between WSI and omics pathway rankings was low at small $k$, exceeded random expectation only at intermediate-to-larger $k$, and remained well below complete agreement (Fig.~\ref{fig:pathway_behavior}d(ii)). Together, these results show that RNA-derived supervision organizes WSI expert usage into pathway-indexed patterns, while the resulting morphological representation captures only a partial view of omics-derived pathway relevance.

\subsection{Human-audited inspection bounds visual interpretation}

\begin{figure}[htbp]
\prfloatbody
\centering
\includegraphics[width=0.96\textwidth]{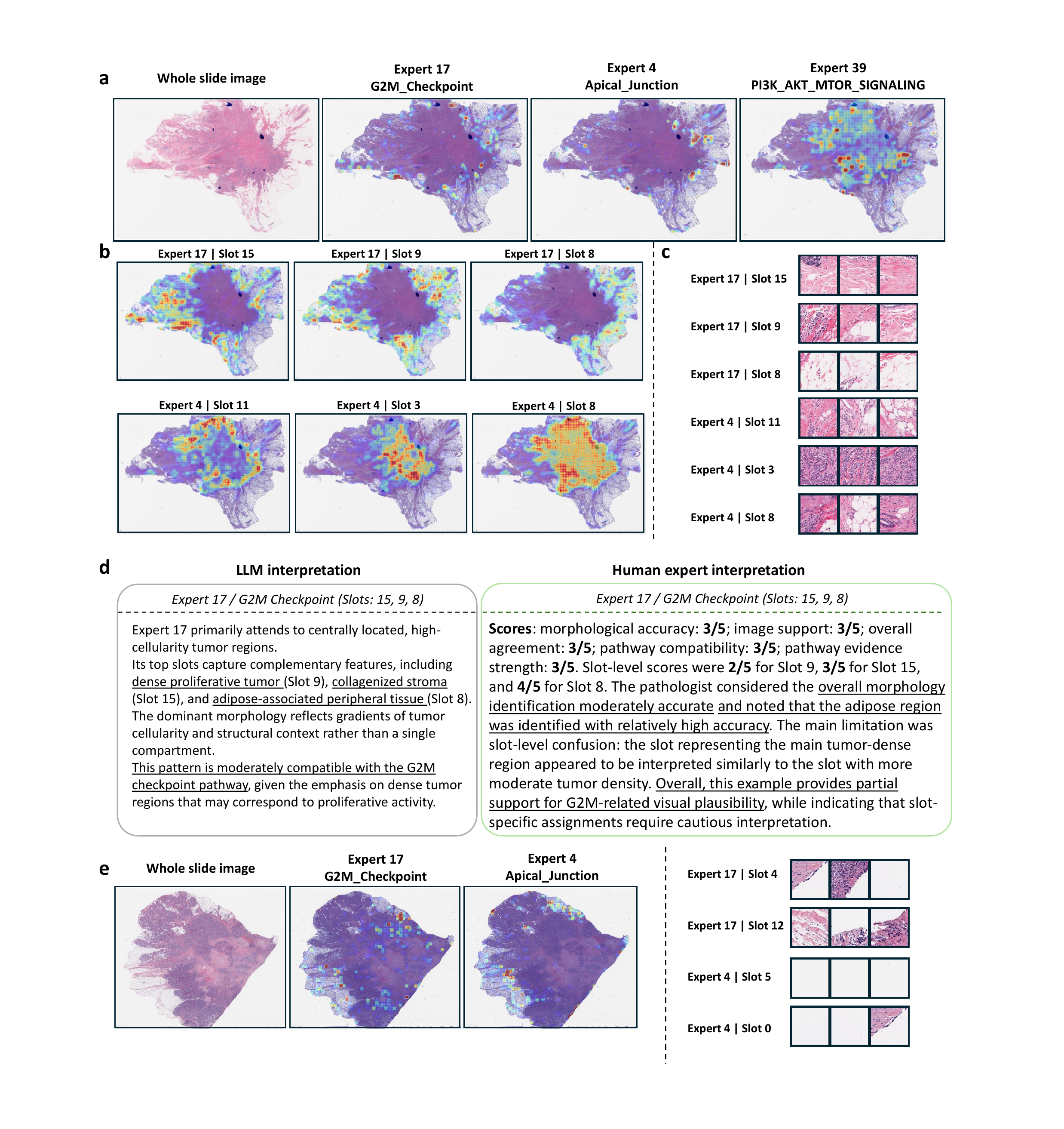}
\caption{
\textbf{Human-audited inspection of pathway-indexed model readouts in BRCA-ODX.}
\textbf{a}, Whole-slide image and attention maps for the three highest-ranked pathway experts.
\textbf{b}, Slot-level attention maps showing distinct spatial patterns within selected experts.
\textbf{c}, Representative high-attention patches for each expert-slot pair.
\textbf{d}, Human-audited interpretation of Expert 17. A language-model morphology summary and corresponding images were reviewed by a senior clinician, providing moderate support for a G2M-compatible morphology hypothesis while revealing slot-level ambiguity.
\textbf{e}, Failure cases with sparse attention or background-rich patches, highlighting sensitivity to preprocessing and patch quality. These visualizations are hypothesis-generating and do not directly validate pathway activation.
}
\label{fig:slot_expert_vis}
\end{figure}

Finally, we examined whether pathway-indexed model behaviour could be inspected through human-reviewable image evidence. Representative correct and failed BRCA-ODX predictions were traced through selected experts, high-gated slots and high-attention patches, followed by a structured expert audit (Fig.~\ref{fig:slot_expert_vis}). A language model was used only to standardize morphology summaries from the model-derived visual evidence, while the clinical expert reviewed both the summaries and source images using predefined criteria for morphological accuracy, image support, pathway compatibility and ambiguity (Supplementary Note~3). This analysis assessed the visual plausibility of pathway-indexed evidence and its failure modes.

In a correctly predicted BRCA-ODX case, the highest-ranked experts corresponded to G2M checkpoint, apical junction and PI3K--AKT--mTOR pathways (Fig.~\ref{fig:slot_expert_vis}a). Slot- and patch-level views revealed the spatial patterns and local morphology underlying expert attention (Fig.~\ref{fig:slot_expert_vis}b,c). For the G2M checkpoint expert, the audit found moderate morphological accuracy, image support and pathway compatibility, while noting slot-level ambiguity, supporting a G2M-compatible morphology hypothesis rather than direct pathway activity (Fig.~\ref{fig:slot_expert_vis}d).

The same workflow also exposed failure modes. In a failed prediction, expert attention was sparse and several high-attention patches corresponded to background or nearly blank tissue regions (Fig.~\ref{fig:slot_expert_vis}e), suggesting degraded evidence due to tissue preprocessing, patch quality or weak morphology. Thus, the inspection chain provides bounded and auditable evidence for where the model attends and where pathway-indexed interpretation may fail.

\section{Discussion}\label{sec:discussion}

MoPE addresses a deployment asymmetry that is common in computational pathology: RNA profiles can inform model development, but routine use often depends on H\&E slides alone. The results are consistent with three observations. First, molecularly supervised histology prediction can be framed as a partial-observability problem rather than only as a conventional missing-modality problem. Second, pathway-structured privileged distillation can improve WSI-only prediction without requiring RNA at inference. Third, pathway-indexed readouts can make model behaviour inspectable, provided they are interpreted as predictive evidence rather than mechanistic proof.

MoPE is motivated by the partial correspondence between transcriptomic activity and tissue morphology: some molecular programs have visible histological correlates, whereas others are only weakly reflected in H\&E. We therefore use pathway structure as biologically organized supervision without assuming that histology can recover the full transcriptomic state. Accordingly, MoPE distills cross-modal information through memory usage rather than direct feature alignment, allowing RNA tokens and pathway-indexed experts to share structured supervision under incomplete cross-modal observability.

Under this formulation, pathway experts can be interpreted as biological-indexed representations of morphology. Consistent with this interpretation, transcriptomic pathway relevance and WSI-side gate separation showed related but non-identical patterns (Fig.~\ref{fig:pathway_behavior}b,c). Such partial correspondence is biologically plausible because molecular risk programs can manifest through histopathological features such as mitotic activity, nuclear grade, and tubule formation \citep{klein2013prediction,geradts2010oncotype}. Thus, pathway expert usage reflects morphological phenotypes associated with molecular programs, rather than direct measurement of their activity.

These pathway readouts provide an auditable intermediate layer between slide-level prediction and local image evidence. Expert usage, slot-level attention and high-attention patches allow predictions to be traced through pathway-indexed representations to specific tissue regions (Fig.~\ref{fig:slot_expert_vis}). However, attention and saliency maps can appear visually plausible without faithfully identifying causal features \citep{adebayo2018sanity,jain2019attention}. Human-audited inspection therefore has a bounded role: it can assess whether highlighted regions support a morphology-linked hypothesis and expose failure modes, but it does not establish molecular activity. Direct validation of the corresponding molecular state would require orthogonal spatial measurements, such as spatial transcriptomics, immunohistochemistry \citep{ZHAO2026114316,andani2025histopathology}, or multiplex protein imaging \citep{HAN2026112458}.

The external evaluation gives this design a more realistic retrospective test. Models trained and selected using public benchmark data were evaluated on two independent institutional cohorts without external RNA input or site-specific refitting for the Oncotype DX test prediction. This setting reflects a common practical constraint: molecular assays may inform model development, whereas image-only inference is often the scalable interface. The external results support further study of pathway-structured privileged distillation for endpoint-specific WSI-only biomarker prediction, particularly for settings in which molecular testing is unavailable. 

Several limitations remain. First, MoPE requires paired WSI--RNA data during training, limiting its development to cohorts with sufficient multimodal annotation. Second, external validation was retrospective and focused primarily on a molecularly informed breast cancer endpoint; broader prospective and multi-institutional evaluation is therefore needed to establish generalizability. Third, model reliability remains sensitive to tissue quality and preprocessing, as illustrated by failure cases involving weak or background-rich regions. Future work should extend validation across institutions and acquisition conditions, incorporate uncertainty-aware quality control, and evaluate pathway-indexed readouts against orthogonal spatial molecular measurements.

\section{Conclusion}\label{sec:conclusion}

In this work, we presented MoPE, a pathway-structured privileged-distillation framework that formulates molecularly supervised computational pathology under partial cross-modal observability. By organizing RNA supervision through Hallmark pathways and transferring it to pathway-indexed histology experts via memory-usage distillation, MoPE improved WSI-only biomarker classification and survival prediction across multiple public cohorts and maintained strong performance in two independent external breast cancer cohorts, while requiring only histology at inference. More broadly, MoPE provides a structured way to exploit molecular information as privileged supervision without assuming full transcriptomic recoverability from morphology, offering a practical framework for deployable and biologically organized computational pathology.

\section*{CRediT authorship contribution statement}
\textbf{Yongxin Guo:} Conceptualization, Methodology, Validation, Visualization, Writing -- original draft, Writing -- review \& editing. \textbf{Hao Lu:} Supervision, Writing -- review \& editing. \textbf{Onur C. Koyun:} Supervision, Writing -- review \& editing. \textbf{Muhammet F. Demir:} Writing -- review \& editing. \textbf{M Khalid Khan Niazi:} Funding acquisition, Supervision, Writing -- review \& editing. \textbf{Metin N. Gurcan:} Conceptualization, Funding acquisition, Supervision, Writing -- review \& editing.

\section*{Declaration of competing interest}
The authors declare that they have no known competing financial interests or personal relationships that could have appeared to influence the work reported in this paper.

\section*{Declaration of generative AI and AI-assisted technologies}
During this study, a large language model (\textit{ChatGPT 5.5}) was used only to convert model-derived visual evidence into a structured morphology summary for subsequent clinical expert review. The output was not used as evidence of pathway activation and was reviewed against the underlying images by a clinical expert. The authors reviewed and edited the content as needed and take full responsibility for the content of the published article.

\section*{Acknowledgments}
We are grateful for support provided by R21 CA273665 (PIs: Gurcan) from the National Cancer Institute, R01 CA276301 (PIs: Niazi, Chen) from the National Cancer Institute, and R21 EB029493 (PIs: Niazi, Segal) from the National Institute of Biomedical Imaging and Bioengineering. The content is solely the responsibility of the authors and does not necessarily represent the official views of the National Institutes of Health, the National Institute of Biomedical Imaging and Bioengineering, or the National Cancer Institute.

\section*{Data availability}
The use of data from The Ohio State University Wexner Medical Center (OSUWMC) was approved by the Ohio State University Cancer Institutional Review Board with a waiver of informed consent and HIPAA authorization. The OSUWMC dataset contains institutional clinical material and is available from the corresponding author upon reasonable request, subject to institutional approval and data-use restrictions. Imaging data from the TCGA project can be accessed at \url{https://portal.gdc.cancer.gov}. The corresponding omics data are available via the UCSC Xena browser at \url{https://xenabrowser.net/datapages/}. The MSigDB 50 Hallmark pathways data can be found at \url{https://www.gsea-msigdb.org}. The Dartmouth Breast Cancer Recurrence Risk Dataset is available from its original source \citep{goyal2024multi}. The source code is uploaded for the review and will be deposited in a public repository upon publication.

\let\bibfont\normalsize
\bibliographystyle{model1-num-names}
\bibliography{references}

\end{document}